\documentclass[10pt,twocolumn,letterpaper]{article}
\pdfoutput=1  

\usepackage[algorithms]{wacv}      

\usepackage{epsfig}
\usepackage{url}
\usepackage{microtype}
\usepackage{seqsplit}
\usepackage{multirow}
\usepackage{algorithm}
\usepackage{algpseudocode}

\def\wacvPaperID{****} 
\def\confName{WACV}
\def\confYear{2027}

\definecolor{wacvblue}{rgb}{0.21,0.49,0.74}
\usepackage[pagebackref,breaklinks,colorlinks,allcolors=wacvblue]{hyperref}

\newcommand{\pp}{\text{pp}}

\begin{document}

\title{The JEPA Predictor:\\A Transferable Operator for Occluded Feature Completion}

\author{%
William Nguyen \qquad Christopher Nguyen\\
Aitomatic, Inc.\\
{\tt\small william@aitomatic.com \quad ctn@aitomatic.com}%
}

\maketitle

\begin{figure*}[t]
\centering
\includegraphics[width=\textwidth]{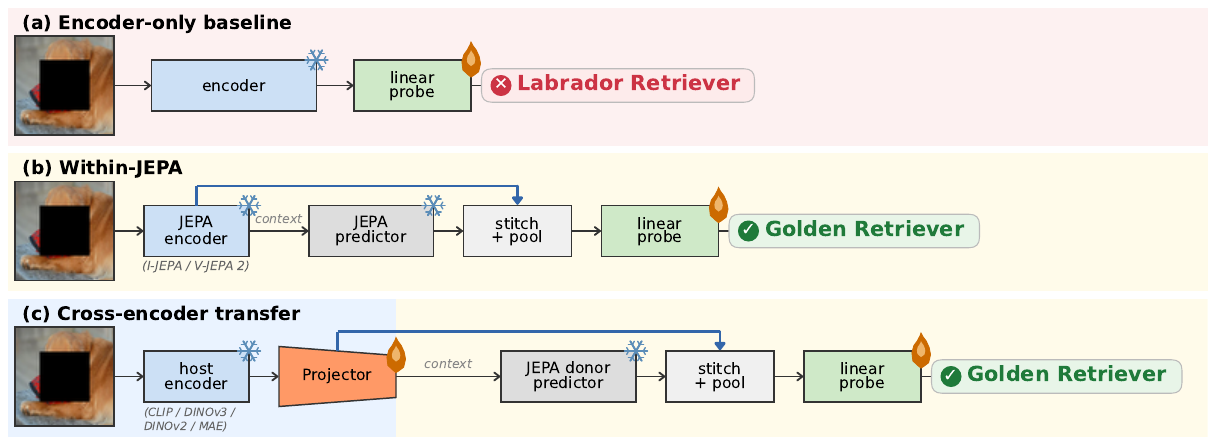}
\caption{\textbf{The protocol's three operating modes (input: real Stanford Dogs image with $K{=}0.25$ center-rect mask applied).} \textbf{(a) Encoder-only baseline:} the predictor is discarded; under heavy occlusion the host encoder alone misclassifies (Labrador) the masked dog. \textbf{(b) Within-JEPA:} a JEPA encoder paired with its own frozen pretraining predictor recovers the prediction (Golden Retriever). The encoder's visible-position features feed both the predictor as context and the downstream stitch (bypass arrow). \textbf{(c) Cross-encoder transfer (this paper's headline configuration):} a non-JEPA host (CLIP, DINOv3, DINOv2, or MAE) is paired with a frozen JEPA donor predictor through a single linear Projector $W$ fit by closed-form ridge least squares on $500$ ImageNet-1k images; the same prediction is recovered without retraining either backbone. Snowflake icons mark frozen modules (encoders and predictors); flame icons mark trainable modules ($W$ is fit once per host-donor pair, the linear probe is fit per cell at the test $K$).}
\label{fig:teaser}
\end{figure*}

\begin{abstract}
Joint-Embedding Predictive Architectures (JEPAs) train a predictor jointly with their encoder, but downstream deployment discards the predictor and reads features from the encoder alone. The predictor is, by construction, a learned operator from visible-context features to features at masked positions, the structure a partial-view classifier needs. We show that this operator is portable across encoder families. We first establish that, at heavy mask, retaining the frozen predictor on a JEPA encoder substantially closes the accuracy gap against the strongest non-JEPA discriminative baselines. We then bolt the frozen predictors of I-JEPA and V-JEPA~2 onto four non-JEPA hosts (CLIP, DINOv3, DINOv2, MAE) through a single linear projection between feature spaces, fit in closed form on 500 ImageNet-1k images. Across both ImageNet-9 and Stanford Dogs and across three mask fractions, the lift over each host's masked-encoder baseline grows monotonically with $K$ in every host--donor pair. CLIP paired with the I-JEPA predictor recovers most of the accuracy that masking removed on ImageNet-9 at heavy occlusion, and lifts fine-grained Stanford Dogs from $15.9\%$ to $52.1\%$ ($+36\pp$). The mechanism is identifiable: the projection pays a fixed cost on visible patches and the predictor provides a growing benefit on masked patches; the benefit dominates the heavy-occlusion regime. At low $K$ on fine-grained classification the projection cost exceeds the benefit, defining the boundary where the linear bridge breaks down. The frozen JEPA predictor functions as a portable operator for occluded feature completion across encoder families, requiring no retraining of either model while fitting matched linear probes per mask fraction.
\end{abstract}

\section{Introduction}
\label{sec:intro}

Frozen image encoders are routinely deployed on partial inputs: truncated medical scans, occluded autonomous-driving frames, interrupted captures. As the mask fraction grows, the strongest publicly available discriminative encoders lose accuracy steeply, and masked-image-modeling and contrastive baselines degrade further still. The natural question is whether a frozen, generic completion module can recover the lost accuracy without retraining the encoder.

Joint-Embedding Predictive Architectures (JEPAs)~\cite{cite-001,cite-003} train a \emph{predictor head} that maps visible-context features and target-position tokens to predicted target features in the encoder's representation space. The downstream protocol typically discards this predictor and reads features from the encoder alone~\cite{cite-001,cite-003,cite-005,cite-008,cite-057,cite-058,cite-064,cite-065,cite-069}. Reusing the predictor at test time recovers accuracy lost to occlusion when the host is itself a JEPA encoder. Whether the predictor's role generalizes beyond the JEPA family is, to our knowledge, untested in the vision literature.

We test that generalization. To pair a frozen JEPA predictor with a non-JEPA host, we fit a single linear projection $W$ between their feature spaces on 500 ImageNet-1k images, in closed form. At test time on a partial input, the host encodes the visible patches, $W$ projects visible-position features into the predictor's input space, and the frozen predictor returns predicted features at the masked positions. A matched linear probe trained at the same mask fraction $K$ reads off the classification. No backbone or predictor parameter is updated.

We evaluate four non-JEPA hosts spanning the major SSL paradigms: contrastive (CLIP~\cite{cite-025}), self-distillation (DINOv3~\cite{cite-023}, DINOv2~\cite{cite-022}), and pixel-reconstruction (MAE~\cite{cite-013}). Each is paired with two donor predictors (I-JEPA~\cite{cite-001}, V-JEPA~2~\cite{cite-003}) at three mask fractions ($K \in \{0.25, 0.56, 0.77\}$), with three seeds, on ImageNet-9~\cite{cite-031,cite-032} and Stanford Dogs~\cite{cite-054}. The lift over each host's masked-encoder baseline grows monotonically with $K$ in every host--donor pair on both datasets. At heavy mask, CLIP paired with the I-JEPA predictor recovers most of its masking-induced accuracy loss on ImageNet-9 and shows its largest lift on Stanford Dogs ($+36\pp$, $15.9\% \to 52.1\%$; Sec.~\ref{sec:results}). The mechanism is identifiable: the projection $W$ pays a fixed cost on visible patches and the predictor provides a growing benefit on masked patches; benefit dominates the heavy-occlusion regime. At low $K$ on fine-grained classification the cost exceeds the benefit, identifying the regime where the linear bridge breaks down.

Two recent works occupy nearby ground. Clin-JEPA~\cite{cite-084} retains the JEPA predictor at inference in a clinical electronic-health-record time-series setting, without a mask-fraction sweep or encoder comparison. CAPI~\cite{cite-085} reports a single-encoder, full-image $k$-NN comparison of encoder-only versus encoder+predictor pooling on a single JEPA model; its predictor outputs are cluster logits, not features, and the evaluation is at a single mask fraction. To our knowledge, no published vision benchmark measures whether a frozen pretraining predictor transfers across encoder families for occluded-feature completion.

\noindent Our contributions are threefold:
\begin{itemize}[leftmargin=*,itemsep=1pt,topsep=2pt]
\item We introduce a protocol that wires a frozen JEPA predictor to any frozen image encoder through a single closed-form linear projection between feature spaces, with no backbone retraining.
\item We benchmark the protocol on four non-JEPA hosts, two JEPA donors, three mask fractions, three seeds, and two datasets, showing that the lift over each host's masked baseline grows monotonically with mask fraction in every host--donor pair.
\item We identify the cost-benefit mechanism that produces the monotone behavior, and the regime where the linear bridge breaks down (low $K$ on fine-grained classification).
\end{itemize}

\section{Related Work}
\label{sec:related}

\noindent\textbf{JEPA family.} I-JEPA~\cite{cite-001,cite-002} trains a predictor to extrapolate target features from context. V-JEPA~2~\cite{cite-003} extends to video; V-JEPA~2-AC~\cite{cite-004} trains a \emph{new} action-conditioned predictor (the pretraining one is not reused). V-JEPA~2.1~\cite{cite-005}, intuitive-physics-from-V-JEPA~\cite{cite-006}, LeJEPA / Gaussian-Embeddings~\cite{cite-007,cite-008}, and JEPA-across-modalities~\cite{cite-009,cite-010,cite-011,cite-012,cite-044,cite-045,cite-046,cite-047,cite-063,cite-064,cite-065,cite-068,cite-069,cite-078} all drop the predictor at downstream-classification inference. Clin-JEPA~\cite{cite-084} first articulated retaining the predictor at inference in clinical EHR time-series, without a $K$-sweep, encoder comparison, or vision evaluation. Independent work~\cite{cite-086} reports a single-point I-JEPA-over-MAE occlusion-robustness gap aligned with our prediction. NEPA~\cite{cite-020} observes qualitatively that next-embedding-prediction representations extrapolate to occluded patches under fine-tuned classification, not frozen-predictor probing.

\noindent\textbf{Masked image modeling.} MAE~\cite{cite-013} and SimMIM~\cite{cite-014} pretrain by pixel-patch reconstruction; the decoder is discarded. Kong and Zhang~\cite{cite-015} argue MIM implicitly learns occlusion invariance. CAPI~\cite{cite-019,cite-085} predicts latent cluster assignments; its Table~5 single-encoder full-image $k$-NN of encoder vs.\ encoder+predictor pooling is the closest single-encoder vision precedent, but its predictor outputs cluster logits rather than features, so it is not reusable as a feature-completion module along a $K$ axis. Surveys~\cite{cite-059}, adaptive-mask variants~\cite{cite-053,cite-060}, the latent-MIM line~\cite{cite-067}, and recent representation-encoders work~\cite{cite-076} broaden the discriminative cohort.

\noindent\textbf{Test-time use of pretraining components.} TTT-MAE~\cite{cite-018} reuses the MAE decoder for gradient adaptation; MAEDAY~\cite{cite-017} for anomaly detection; CuMoLoS-MAE~\cite{cite-016} with mask-ratio sweeps; Continual-MAE~\cite{cite-061}, MATE~\cite{cite-062}, ETA~\cite{cite-070} for continual / 3D / depth. Masksembles~\cite{cite-041} uses inference-time masks for ensembling. V-JEPA~2-AC~\cite{cite-004} is the closest JEPA precedent but trains a new predictor. Our cell (frozen encoder, frozen pretraining predictor, forward-only, classification) is unoccupied.

\noindent\textbf{Partial-view and occlusion classification.} FCFormer~\cite{cite-039}, Region-Feature-Completion~\cite{cite-040}, Deep-feature-augmentation~\cite{cite-075} train \emph{task-specific} completion modules for occluded Re-ID; ViT-SSL-for-Re-ID~\cite{cite-080} confirms mask ratios near $0.50$ are informative. Im2Flow~\cite{cite-042}, ImageNet-E~\cite{cite-074}, foreground-background~\cite{cite-031,cite-033}, Masked-Adaptive-Ensemble~\cite{cite-077}, MIMIR~\cite{cite-079} provide adjacent benchmarks. Garrido \etal~\cite{cite-006} use V-JEPA prediction error as a violation-of-expectation signal in physics. The distinguishing feature here is a \emph{pretrained, frozen, generic} predictor rather than a task-specific module.

\noindent\textbf{Frozen-encoder probing.} DINOv2~\cite{cite-022} and DINOv3~\cite{cite-023} established frozen-encoder probing as the default self-supervised-learning (SSL) evaluation; iBOT~\cite{cite-024}, DINO~\cite{cite-021} are antecedents. Battle of the Backbones~\cite{cite-035}, Lexicon3D~\cite{cite-030}, Probing 3D Awareness~\cite{cite-036}, Probing Mid-Level Vision~\cite{cite-034}, Attention-Please~\cite{cite-038} (our attentive-probe precedent), and RankMe~\cite{cite-037} extend the methodology. CLIP-family backbones~\cite{cite-025,cite-026,cite-027,cite-028,cite-029,cite-072} complete the comparator set. SimSiam~\cite{cite-057}, BYOL~\cite{cite-058}, and the Apple implicit-bias analysis~\cite{cite-066} document that predictor heads are conventionally discarded post-training.

\noindent\textbf{Cross-encoder feature-space alignment.} A long line of work maps features between independently trained encoder spaces. Lenc and Vedaldi~\cite{cite-087} introduced linear stitching to test representation equivalence by fitting a single linear layer between an early stage of one CNN and a later stage of another. Bansal \etal~\cite{cite-088} revisited this idea as a tool to compare neural representations, showing that small linear stitching layers between two frozen networks recover most of the source network's downstream task accuracy. Moschella \etal~\cite{cite-089} encode each sample by its cosine similarity to a fixed set of anchor samples, producing an anchor-frame representation that is invariant to encoder-specific rotations and enables zero-shot communication between independently trained networks. Git Re-Basin~\cite{cite-090} aligns independently trained networks in weight space by exploiting the permutation symmetry of neural networks. The cross-encoder bridge $W$ used here is a closed-form ridge-fit instance of this stitching family, with three distinguishing features. First, the alignment is between two \emph{image encoders} of different families (CLIP, DINOv3, DINOv2, MAE as hosts; I-JEPA or V-JEPA~2 as donor target) rather than between layers of one family or between two CNNs trained on the same task. Second, the downstream task is not classification stitching but \emph{feature completion}: the donor's frozen predictor consumes $W$-projected host features at visible positions and returns predicted features at masked positions, which are then classified by a matched probe. Third, $W$ is fit on $500$ unmasked image pairs and reused unchanged across all mask fractions, datasets, and seeds; no trainable parameters are added downstream. The model-stitching literature establishes that a single linear layer between two frozen encoders is a defensible baseline; this paper inherits that defense and uses it to study a different downstream operator (the JEPA predictor as a transferable completion operator).

\section{Method}
\label{sec:method}

\noindent\textbf{Notation.} $K \in [0, 1]$: mask fraction (the proportion of patches masked). $D'(K)$: matched-probe accuracy of an encoder alone on a $K$-masked image (the no-predictor baseline). $D(K)$: matched-probe accuracy of the same protocol with the predictor's completion plugged in. $\Delta(K) \equiv D(K) - D'(K)$: the per-cell lift in percentage points (pp). $W$: the linear projection bridge between a non-JEPA host's feature space and a donor JEPA encoder's feature space (Sec.~\ref{sec:method}, Cross-encoder bridge). NN: a nearest-neighbor patch-completion baseline that fills each masked patch with the mean of its 5 cosine-nearest visible-patch features (Sec.~\ref{sec:method}, Baselines). Top-1 accuracy is reported in \%; lifts in pp.

\noindent\textbf{JEPA setup.} A JEPA model~\cite{cite-001,cite-003} has a context encoder $f_c$, target encoder $f_t$, and predictor $\pi$. Pretraining minimizes $\sum_i \| \pi(f_c(X_c), \text{pos}(X_{t,i})) - \text{sg}(f_t(X_{t,i})) \|^2$ over context/target partitions. At convergence the predictor is a learned extrapolation operator from (visible-context features, target-position) to predicted target features. Conventionally it is discarded after pretraining; $f_t$ (or its exponential moving average (EMA) stable copy) is the production encoder.

\noindent\textbf{Test-time protocol.} Let $x$ be an image and $M \in \{0,1\}^{P}$ a patch-level mask ($M_p = 1$ marks a masked patch); $K = \frac{1}{P}\sum_p M_p$. Define $F_{\text{partial}} = f(x \odot (1-\text{up}(M)))$ (encode the partial image; post-norm zero-fill at the patch-token interface~\cite{cite-053}); $F_{\text{predicted}} = \pi(F_{\text{partial}}, \text{pos}(M))$; $F_{\text{completed}} = \text{stitch}(F_{\text{partial}}, F_{\text{predicted}}; M)$ (length-$P$: masked positions take predicted features, visible take encoded); $F_{\text{full}} = f(x)$ (oracle only). A linear probe is trained on mean-pooled $F_{\text{completed}}$ at the same $K$ used at test (a \emph{matched probe}); we write this $\text{acc}_{\text{completed\_matched}}(K)$.

\noindent\textbf{Why matched probing.} The unmatched alternative (a probe fit on $F_{\text{full}}$ and evaluated on $F_{\text{completed}}$) collapses to near-chance in pilot measurements: $F_{\text{completed}}$ is structurally different from $F_{\text{full}}$ (half its components are predictor outputs), so a classifier fit on full-image features has no reason to generalize. Matched probing is well-established in adjacent literatures (TTT-MAE~\cite{cite-018} re-fits per-sample; occluded Re-ID~\cite{cite-039,cite-040} trains on occluded features) and is cheap (per-$K$ linear fit over frozen features; minutes on a single GPU). It applies symmetrically to the predictor cell and to the encoder-only baseline. We treat $K$ as known at inference, so the protocol uses a contiguous train/val partition per $K$.

\noindent\textbf{V-JEPA 2's frame-shifted predictor.} A natural consistency check asks whether $F_{\text{predicted}}$ is close to $F_{\text{full}}$. For I-JEPA, predictor outputs are meaningfully closer to full-image features than a trivial mean baseline under relative L2 reconstruction error. For V-JEPA~2, the same L2 criterion would flag the predictor's outputs as uninformative, yet matched probing recovers strong accuracy (Sec.~\ref{sec:results}). The predictor's outputs are classification-useful but live in a different L2 frame than the encoder's. We adopt matched-probe accuracy, rather than L2 reconstruction error, as the appropriate diagnostic for any reused JEPA predictor whose pretraining loss does not constrain its outputs to the target encoder's metric.

\noindent\textbf{Cross-encoder bridge.} For non-JEPA hosts the predictor's input is foreign: the host's features live in a different space than the donor JEPA encoder's. A linear projection $W : \mathbb{R}^{D_\text{host}} \to \mathbb{R}^{D_\text{JEPA}}$ is fit on $500$ ImageNet-1k images by closed-form ridge least squares ($\lambda{=}10^{-2}$), minimizing per-token MSE between host features and donor encoder features at matched patch positions. $W$ is fit once per (host, donor) pair on unmasked images and reused across all $K$, datasets, and seeds. At test time on a partial image, the host encodes the visible patches; $W$ projects visible-position features into the donor's space; the frozen donor predictor returns predicted features at masked positions; the stitched sequence lives in the donor's embedding space and is classified by a matched probe trained at the same $K$. No backbone or predictor parameter is updated. \Cref{alg:protocol} summarizes the test-time protocol.

\begin{algorithm}[t]
\caption{Cross-encoder latent extrapolation at test time. Within-JEPA is the special case $W{=}I$ (no bridge).}
\label{alg:protocol}
\small
\begin{algorithmic}[1]
\Require partial image $x$, patch mask $M$ ($K{=}\tfrac{1}{P}\sum_p M_p$); frozen host encoder $f$; frozen donor predictor $\pi$; bridge $W$ (fit once per host--donor pair); matched probe $g_K$
\State $F_{\text{partial}} \gets f\big(x \odot (1-\mathrm{up}(M))\big)$ \Comment{encode visible patches}
\State $H \gets W\,F_{\text{partial}}$ \Comment{project into donor space}
\State $F_{\text{pred}} \gets \pi\big(H,\ \mathrm{pos}(M)\big)$ \Comment{frozen predictor completes masked positions}
\State $F_{\text{comp}} \gets \mathrm{stitch}(H, F_{\text{pred}}; M)$ \Comment{visible$\gets H$, masked$\gets F_{\text{pred}}$}
\State \textbf{return} $g_K\big(\mathrm{meanpool}(F_{\text{comp}})\big)$ \Comment{matched linear probe at test $K$}
\end{algorithmic}
\end{algorithm}

\noindent\textbf{Computational cost.} The protocol trains neither backbone. The bridge $W$ is a single $D_\text{host}\times D_\text{JEPA}$ matrix ($1024\times\{1280, 1024\}$, ${\sim}1.0$--$1.3$M parameters) from one closed-form ridge solve on $500$ images (a few seconds on one GPU), fit once per (host, donor) pair and reused across all $K$, datasets, and seeds. Inference is one frozen host forward pass plus one frozen predictor forward pass; the only component fit at test time is the linear matched probe (one per $K$, seconds over cached features). No gradient flows through the host encoder or the donor predictor at any stage.

\noindent\textbf{Baselines (5 per cell).} (1)~$\text{acc}_{\text{full}}$, oracle on unmasked image. (2)~$\text{acc}_{\text{partial\_matched}}\ (D')$, fair encoder-only floor: matched probe trained on $F_{\text{partial}}$. (3)~$\text{acc}_{\text{completed\_matched}}\ (D)$, predictor protocol. (4)~$\text{acc}_{\text{mae\_pixel\_matched}}$: MAE ViT-L decoder fills masked patches in pixel space; the completed image is re-encoded; matched probe is fit. (5)~$\text{acc}_{\text{nearest\_neighbor\_completion}}$: fill each masked patch's slot with the cosine-similarity mean of 5 nearest visible-patch features. Lifts: $\Delta_{\text{pred}} \equiv D - D'$; $\Delta_{\text{vs\_mae}} \equiv D - \text{acc}_{\text{mae\_pixel\_matched}}$; $\Delta_{\text{vs\_NN}} \equiv D - \text{acc}_{\text{nearest\_neighbor\_completion}}$. Per-cell CIs use paired bootstrap (1000) on lifts~\cite{cite-082} and Wilson 95\% on accuracies~\cite{cite-081}. Across-seed aggregation reports Welch-$t$ 95\% at df${=}2$~\cite{cite-083} (the $t_2 \approx 4.303$ factor inflates intervals conservatively, the appropriate stance for $n{=}3$).

\section{Experimental Setup}
\label{sec:setup}

\noindent\textbf{Datasets.} Headline: ImageNet-9~\cite{cite-031,cite-032}, 4050 images, 9-way coarse classification. Generalization: (1)~Stanford Dogs~\cite{cite-054} (120-way fine-grained), 9600 images stratified ($80\text{/class} \times 120$), 80/20-split into 7680 train / 1920 val per matched-probe cell; (2)~ImageNet-1k val~\cite{cite-056} (1000-way), 9000 images stratified ($9\text{/class} \times 1000$), 70/30-split into 7000 train / 2000 val per matched-probe cell.

\noindent\textbf{Encoders.} Headline (6): I-JEPA~\cite{cite-001}, V-JEPA~2~\cite{cite-003}, DINOv3~\cite{cite-023}, DINOv2~\cite{cite-022}, CLIP~\cite{cite-025}, MAE~\cite{cite-013}. Appendix at $K{=}0.56$ only: SigLIP~\cite{cite-026}, MetaCLIP~\cite{cite-028} (OpenCLIP~\cite{cite-029}). Predictor cells (JEPA only): vendored pretraining predictors; checkpoint SHA-256 hashes in the reproducibility appendix.

\noindent\textbf{Mask fractions ($K$ = fraction of masked patches).} Within-JEPA sweep uses $K \in \{0.14, 0.25, 0.56, 0.66, 0.77\}$, patch-counted as $\{36, 64, 144, 169, 196\}/256$ on the $16{\times}16$ patch grid (center-rect masks of side $\{6, 8, 12, 13, 14\}$ patches). The center-rect builder caps at $K{=}0.77$ (a $14{\times}14$ mask, the largest center-aligned rectangle with a one-patch margin). Cross-encoder sweep (Sec.~\ref{sec:cross-encoder}) uses the three-point subset $K \in \{0.25, 0.56, 0.77\}$.

\noindent\textbf{Mask shapes.} Center-rect (headline), off-center, connected blob, random Bernoulli. \textbf{Probe class.} Linear (headline) and a 4-layer attentive probe~\cite{cite-038}. \textbf{Seeds.} $\{0, 1, 2\}$. Encoder-scale and iterative-refinement ablations were dropped per compute budget; see Sec.~\ref{sec:limitations}. Code entrypoint, per-cell artifact layout, and aggregation paths are documented in the Reproducibility Appendix.

\section{Results}
\label{sec:results}

The section is structured to (i) state the headline cross-encoder finding (Sec.~\ref{sec:cross-encoder}), (ii) explain the cost-benefit mechanism that drives it (Sec.~\ref{sec:mechanism}), (iii) report the within-JEPA specialization and rule out three alternative explanations of the lift (Sec.~\ref{sec:within-jepa}), and (iv) test generalization on Stanford Dogs and ImageNet-1k val (Sec.~\ref{sec:transfer}). All accuracies are top-1 (\%) over 3 seeds; cross-seed CIs use Welch-$t$ $95\%$ at df${=}2$.

\subsection{Cross-encoder transfer is the headline finding}
\label{sec:cross-encoder}

\begin{figure}[t]
\centering
\includegraphics[width=\linewidth]{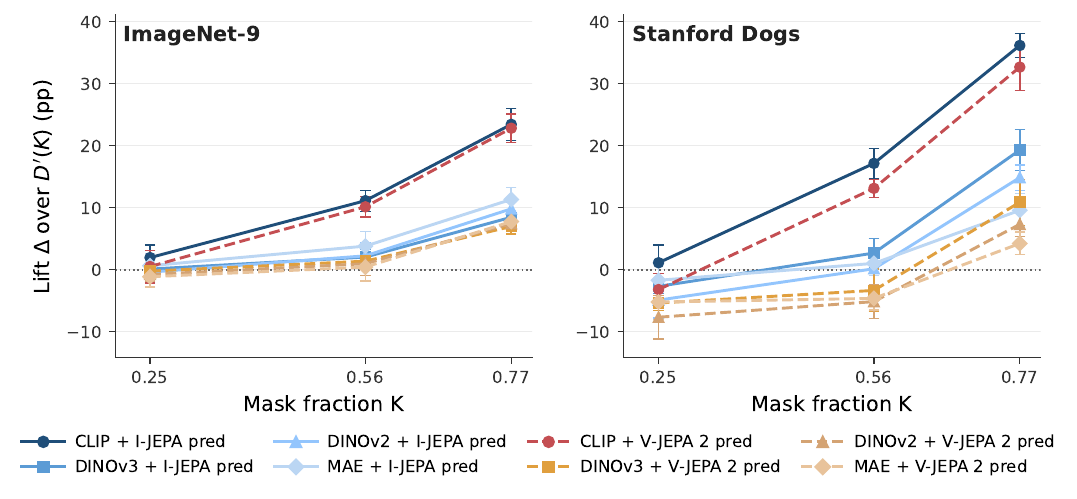}
\caption{Per-pair lift $\Delta = D - D'$ over each host's matched baseline as a function of mask fraction $K$. Left: IN-9. Right: Stanford Dogs. Solid: I-JEPA donor; dashed: V-JEPA~2 donor. Marker shape distinguishes hosts. Error bars: Welch-$t$ $95\%$ CI. The lift is monotone in $K$ in every pair and goes negative at low $K$ on Stanford Dogs for several pairs (cost-benefit boundary).}
\label{fig:cross-encoder}
\end{figure}

The lift over each host's $D'(K)$ baseline grows monotonically with $K$ in every host--donor pair on both ImageNet-9 and Stanford Dogs (\Cref{tab:main-results}, \Cref{fig:cross-encoder}). At $K{=}0.77$, $7$ of the $8$ host--donor pairs (4 hosts $\times$ 2 donors) show a positive lift with Welch-$t$ $95\%$ CI strictly above zero on both datasets; the eighth (MAE+V-JEPA~2 on Stanford Dogs, $+4.25\pp$) is positive in all three seeds but its CI just clears zero on one dataset only. For each cell, the host encodes once, $W$ projects, the frozen donor predictor runs once, and the matched linear probe is fit on the completed sequence.

\begin{table*}[t]
\centering
\footnotesize
\begin{tabular*}{\textwidth}{@{\extracolsep{\fill}}lcccccc}
\toprule
& \multicolumn{3}{c}{ImageNet-9 (9-way)} & \multicolumn{3}{c}{Stanford Dogs (120-way)} \\
\cmidrule(lr){2-4} \cmidrule(lr){5-7}
Method & $K{=}0.25$ & $K{=}0.56$ & $K{=}0.77$ & $K{=}0.25$ & $K{=}0.56$ & $K{=}0.77$ \\
\midrule
\multicolumn{7}{@{}l}{\emph{Non-JEPA host paired with frozen JEPA predictor through bridge $W$ (cross-encoder transfer):}} \\
CLIP                  & $92.72$ & $85.06$ & $73.62$ & $50.64$ & $32.95$ & $15.87$ \\
\quad + I-JEPA pred   & $94.70$\,{\tiny(+1.98)}  & $96.21$\,{\tiny(+11.15)} & $97.08$\,{\tiny(+23.46)} & $51.79$\,{\tiny(+1.15)}  & $50.12$\,{\tiny(+17.17)} & $52.05$\,{\tiny(+36.18)} \\
\quad + V-JEPA~2 pred & $93.25$\,{\tiny(+0.53)}  & $95.22$\,{\tiny(+10.16)} & $96.46$\,{\tiny(+22.84)} & $47.46$\,{\tiny($-$3.18)} & $46.09$\,{\tiny(+13.14)} & $48.56$\,{\tiny(+32.69)} \\
\midrule
DINOv3                & $97.82$ & $95.31$ & $88.60$ & $\mathbf{72.40}$ & $58.56$ & $38.98$ \\
\quad + I-JEPA pred   & $\mathbf{97.94}$\,{\tiny(+0.12)} & $97.37$\,{\tiny(+2.06)} & $97.08$\,{\tiny(+8.48)} & $69.74$\,{\tiny($-$2.66)} & $\mathbf{61.27}$\,{\tiny(+2.71)} & $\mathbf{58.30}$\,{\tiny(+19.32)} \\
\quad + V-JEPA~2 pred & $97.61$\,{\tiny($-$0.21)} & $96.71$\,{\tiny(+1.40)} & $95.60$\,{\tiny(+7.00)} & $67.05$\,{\tiny($-$5.35)} & $55.24$\,{\tiny($-$3.32)} & $49.88$\,{\tiny(+10.90)} \\
\midrule
DINOv2-ViT-L/14       & $97.53$ & $94.28$ & $86.30$ & $65.16$ & $50.17$ & $31.72$ \\
\quad + I-JEPA pred   & $97.16$\,{\tiny($-$0.37)} & $96.54$\,{\tiny(+2.26)} & $96.14$\,{\tiny(+9.84)} & $60.28$\,{\tiny($-$4.88)} & $50.34$\,{\tiny(+0.17)} & $46.60$\,{\tiny(+14.88)} \\
\quad + V-JEPA~2 pred & $96.79$\,{\tiny($-$0.74)} & $95.27$\,{\tiny(+0.99)} & $93.75$\,{\tiny(+7.45)} & $57.52$\,{\tiny($-$7.64)} & $45.05$\,{\tiny($-$5.12)} & $39.06$\,{\tiny(+7.34)} \\
\midrule
MAE                   & $88.97$ & $81.81$ & $72.26$ & $47.93$ & $30.92$ & $15.59$ \\
\quad + I-JEPA pred   & $89.63$\,{\tiny(+0.66)} & $85.64$\,{\tiny(+3.83)} & $83.62$\,{\tiny(+11.36)} & $46.25$\,{\tiny($-$1.68)} & $31.96$\,{\tiny(+1.04)} & $25.19$\,{\tiny(+9.60)} \\
\quad + V-JEPA~2 pred & $87.86$\,{\tiny($-$1.11)} & $82.26$\,{\tiny(+0.45)} & $80.04$\,{\tiny(+7.78)} & $42.74$\,{\tiny($-$5.19)} & $26.32$\,{\tiny($-$4.60)} & $19.84$\,{\tiny(+4.25)} \\
\midrule[\heavyrulewidth]
\multicolumn{7}{@{}l}{\emph{JEPA encoder paired with its own frozen native predictor (within-JEPA, no bridge):}} \\
I-JEPA                & $95.68$ & $89.59$ & $75.88$ & $44.08$ & $32.12$ & $14.64$ \\
\quad + own predictor & $97.28$\,{\tiny(+1.60)} & $\mathbf{97.78}$\,{\tiny(+8.19)} & $\mathbf{98.11}$\,{\tiny(+22.23)} & $46.75$\,{\tiny(+2.67)} & $43.40$\,{\tiny(+11.28)} & $40.68$\,{\tiny(+26.04)} \\
\midrule
V-JEPA~2              & $95.39$ & $91.32$ & $82.18$ & $40.38$ & $32.38$ & $18.92$ \\
\quad + own predictor & $95.97$\,{\tiny(+0.58)} & $96.87$\,{\tiny(+5.55)} & $97.57$\,{\tiny(+15.39)} & $43.02$\,{\tiny(+2.64)} & $39.43$\,{\tiny(+7.05)} & $37.43$\,{\tiny(+18.51)} \\
\bottomrule
\end{tabular*}
\caption{Matched-probe top-1 accuracy (\%) on IN-9 and Stanford Dogs across mask fractions. \emph{Top block (non-JEPA hosts):} CLIP, DINOv3, DINOv2, MAE with and without a frozen JEPA donor predictor bolted on through a single linear bridge $W$ fit on $500$ ImageNet-1k images. \emph{Bottom block (JEPA hosts):} I-JEPA and V-JEPA~2 with and without their own native frozen predictor (no bridge). Thin rules separate each host's $\{$without, +I-JEPA, +V-JEPA~2$\}$ triple; the heavy rule separates cross-encoder from within-JEPA. \textbf{Bold} = best per column. At $K{=}0.77$ all predictor-augmented rows exceed their corresponding encoder-only rows; all within-JEPA rows and $7$ of $8$ cross-encoder host--donor pairs have Welch-$t$ $95\%$ CI above zero on both datasets, while MAE+V-JEPA~2 on Stanford Dogs is positive in all three seeds. At $K{=}0.25$ on Stanford Dogs, several host--donor pairs go negative (cost of $W$ exceeds predictor benefit on fine-grained classification, the cost-benefit boundary discussed in Sec.~\ref{sec:cross-encoder}). Per-pair $\Delta$ curves with CIs in \Cref{fig:cross-encoder}.}
\label{tab:main-results}
\end{table*}

\begin{figure}[t]
\centering
\includegraphics[width=\linewidth]{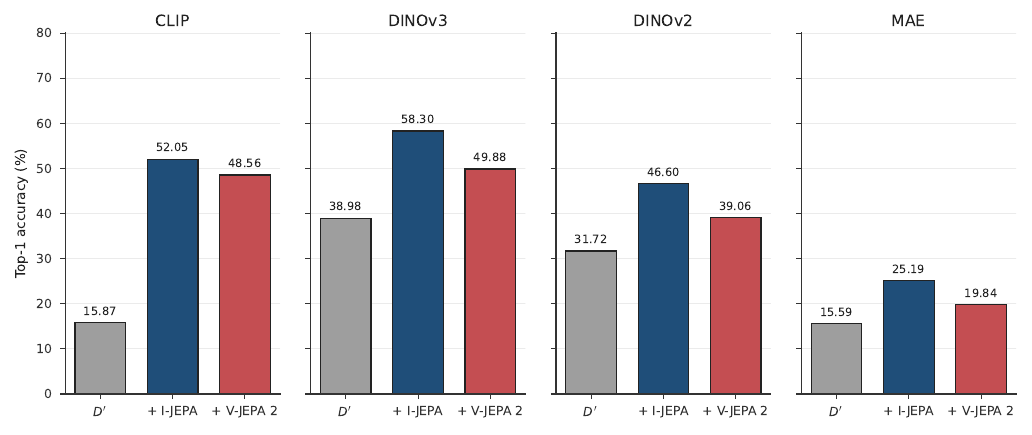}
\caption{Matched-probe top-1 accuracy (\%) on Stanford Dogs at $K{=}0.77$ for each host alone ($D'$) and the same host paired with each frozen JEPA donor predictor ($D$). 3-seed means. The lift over $D'$ is largest where $D'$ is smallest (CLIP).}
\label{fig:per-host}
\end{figure}

The pattern is sharpest on CLIP, whose contrastive caption-alignment objective does not explicitly train for local image-feature structure. CLIP loses the most accuracy under masking ($D'$ falls $92.72\% \to 73.62\%$ across $K{\in}[0.25, 0.77]$ on ImageNet-9; $50.64\% \to 15.87\%$ on Stanford Dogs), and the frozen JEPA predictor recovers most of the IN-9 gap: CLIP+I-JEPA reaches $+23.46\pp$ at $K{=}0.77$ on ImageNet-9 (of the ${\sim}25\pp$ masking removed) and $+36.18\pp$ on Stanford Dogs. DINOv3 and DINOv2, whose self-distillation objectives already build occlusion-robust local features, show smaller but still monotonically growing lifts (\Cref{tab:main-results}). MAE, the generative pixel-reconstruction host, behaves in between: its lower base accuracy leaves headroom comparable to CLIP, but its features sit at a less favourable angle to the JEPA input manifold and the recovered lifts are roughly half of CLIP's ($+11.36\pp$ on IN-9 and $+9.60\pp$ on Stanford Dogs at $K{=}0.77$ with I-JEPA). The I-JEPA donor outperforms the V-JEPA~2 donor in every host pair, consistent with I-JEPA's image-only pretraining matching the still-image evaluation domain.

\subsection{Mechanism: a cost-benefit balance with a pretraining-mask-ratio peak}
\label{sec:mechanism}

The $K$-dependent shape of the lift, the negative entries at low $K$ on fine-grained Stanford Dogs, and the location of the peak all follow from a single cost-benefit account.

\noindent\textbf{Cost on visible patches, benefit on masked patches.} The projection $W$ pays a cost on \emph{all} visible patches: its ridge fit explains only a fraction of donor-feature variance (fit-set $R^2_W$ from $0.147$ for CLIP to $0.292$ for DINOv3 with the I-JEPA donor; \Cref{tab:bridge}), so $W$ is a useful but lossy bridge. This $R^2_W$ measures \emph{projection cost}, not success: as with the V-JEPA~2 frame shift (Sec.~\ref{sec:method}), feature-space fidelity does not track task usefulness, and the bridge's validity rests on the W-only control below and the predictor-specific, mask-monotone lift, not on feature reconstruction. Across the four hosts $R^2_W$ is inversely ordered with the lift (\Cref{tab:bridge}): the lowest-$R^2_W$ host (CLIP) shows the largest $\Delta$ and the highest (DINOv3) the smallest. This four-host ordering is qualitative, but matches the cost-benefit account: a host whose features are hardest to align linearly is also one whose native features are weakest under occlusion, so it pays a larger projection cost yet the predictor's benefit dominates by a wider margin. The predictor's completion is injected only at \emph{masked} patches. At $K{=}0.25$ the cost is paid on $\sim 192$ visible tokens and the benefit accrues on $\sim 64$ masked tokens; at $K{=}0.77$ the balance inverts, with $\sim 64$ visible and $\sim 192$ masked. The observed net lift becomes positive when masked-position benefit overtakes visible-position projection cost. On fine-grained $120$-way Stanford Dogs the projection cost is larger because fine-grained discriminability depends on subtle feature components $W$ cannot fully preserve, so the cost-benefit boundary shifts to a larger $K$ than on coarser $9$-way ImageNet-9. This is the negative-lift region visible at $K{=}0.25$ on Stanford Dogs in \Cref{tab:main-results} and \Cref{fig:cross-encoder}.

\noindent\textbf{The bridge is a faithful but lossy translator.} The W-only sanity baseline (apply $W$ at all positions, no predictor, then probe) sits at or just below each host's own $D'(K)$ at every measured $K$ (\Cref{tab:bridge}): at $K{=}0.77$ on IN-9, CLIP's W-only accuracy is $71.6\%$ against encoder-only $D'{=}73.6\%$, while the full protocol reaches $D{=}97.1\%$. A random or degenerate projection would drive W-only toward chance; instead $W$ preserves the host's discriminative structure on visible patches, and the predictor's masked-position output supplies the entire lift. The projection is a translator, not a content generator. The W-only cost is moreover small in absolute terms: at $K{=}0.77$ W-only trails $D'$ by at most ${\sim}3\pp$ (DINOv2 $-0.1$, DINOv3 $-0.6$, CLIP $-2.0$, MAE $-3.2$), so the projection retains almost all of the host's visible-patch discriminability and the large net lifts come overwhelmingly from the predictor's completion, not from any change $W$ makes to visible features.

\begin{table}[t]
\centering
\small
\setlength{\tabcolsep}{4pt}
\begin{tabular}{lccccc}
\toprule
Host & $R^2_W$ & $D'$ & W-only & $D$ & $\Delta$ \\
\midrule
CLIP    & $0.147$ & $73.6$ & $71.6$ & $\mathbf{97.1}$ & $\mathbf{+23.5}$ \\
MAE     & $0.229$ & $72.3$ & $69.1$ & $83.6$ & $+11.4$ \\
DINOv2  & $0.232$ & $86.3$ & $86.2$ & $96.1$ & $+9.8$ \\
DINOv3  & $0.292$ & $88.6$ & $88.0$ & $97.1$ & $+8.5$ \\
\bottomrule
\end{tabular}
\caption{Bridge quality vs.\ lift on IN-9 at $K{=}0.77$ with the I-JEPA donor (hosts ordered by ascending bridge fit $R^2_W$). $R^2_W$: fraction of donor-feature variance the ridge bridge $W$ explains on its $500$-image fit set, a \emph{cost}/lossiness measure, not accuracy. $D'$: encoder-only. \mbox{W-only}: $W$ applied at all positions, no predictor. $D$: full protocol. $\Delta{=}D{-}D'$. W-only sits at or below $D'$ for every host (the bridge preserves but does not add) while $D$ supplies the lift; $R^2_W$ is inversely ordered with $\Delta$.}
\label{tab:bridge}
\end{table}

\noindent\textbf{Why the lift peaks near the pretraining mask ratio.} The predictor was trained to predict masked-region features from visible-region features at a fixed mask-ratio distribution; its operating point is therefore set by pretraining. At small $K$ the encoder alone retains most of the relevant context, so the predictor has little marginal information to add. As $K$ grows the encoder loses context, and the predictor's contribution grows accordingly. I-JEPA peaks at $K{=}0.77$ in our grid, near its pretraining mask ratio (effective $\approx 75\%$~\cite{cite-001}). V-JEPA~2 also peaks at observed $K{=}0.77$ in our grid; its pretraining mask ratio is closer to $90\%$~\cite{cite-003}, but the center-rect mask builder saturates at actual $K{=}0.77$ on the $16{\times}16$ patch grid (Sec.~\ref{sec:setup}), so V-JEPA~2's peak at its true pretraining ratio cannot be directly tested with this geometry.

\noindent\textbf{The pattern holds for both donors.} The V-JEPA~2 donor reproduces every qualitative feature of the I-JEPA donor on IN-9: W-only stays at or below $D'$ at all measured $K$, the lift grows monotonically with $K$ in every host, and the bridge $R^2_W$ (lower for V-JEPA~2, $0.137$--$0.241$, since its predictor lives in a different feature frame; Sec.~\ref{sec:method}) is again inversely ordered with the lift across hosts. The V-JEPA~2 lifts are uniformly smaller than I-JEPA's at $K{=}0.77$ (CLIP $+22.8$ vs.\ $+23.5\pp$; DINOv3 $+7.0$ vs.\ $+8.5\pp$; DINOv2 $+7.4$ vs.\ $+9.8\pp$; MAE $+7.8$ vs.\ $+11.4\pp$), consistent with I-JEPA's image-only pretraining better matching the still-image evaluation domain. That the same cost-benefit ordering survives a donor whose predictor sits in a different feature frame is further evidence that the transfer is a property of the predictor-as-completion-operator, not of any one feature geometry.

The same host-strength-vs-lift ordering recurs on fine-grained Stanford Dogs at $K{=}0.77$ (\Cref{tab:main-results}): CLIP, the weakest host under masking, again shows the largest I-JEPA-donor lift ($+36.2\pp$), while the self-distillation hosts DINOv3 and DINOv2, which already carry occlusion-robust local structure, show the smaller lifts ($+19.3$ and $+14.9\pp$). The ordering is therefore a property of host--predictor complementarity, not a coarse-task artifact of $9$-way IN-9.
\subsection{Within-JEPA specialization and three mechanism checks}
\label{sec:within-jepa}

The native specialization of the protocol (a JEPA encoder paired with its own predictor, with no cross-encoder bridge) removes the projection cost entirely and provides the cleanest empirical handle on the predictor's contribution. I-JEPA + frozen predictor maintains $97.78$--$98.11\%$ top-1 accuracy across $K \in [0.56, 0.77]$, while every encoder-only baseline drops steeply: DINOv3 from $98.60\%$ at $K{=}0.14$ to $88.60\%$ at $K{=}0.77$; DINOv2-ViT-L/14@224 from $98.40\%$ to $86.30\%$; CLIP and MAE end at $73.62\%$ and $72.26\%$ respectively at $K{=}0.77$ (\Cref{tab:main-results}, bottom block). At $K{=}0.77$, I-JEPA + predictor reaches $\boldsymbol{98.11\%}$, exceeding the strongest encoder-only baseline (DINOv3) by $\boldsymbol{+9.51\pp}$ even though DINOv3 leads at every other $K$. The within-JEPA curves cross the encoder-only curves between $K{=}0.25$ and $K{=}0.56$.

Three mechanism checks rule out alternative explanations of the within-JEPA lift (\Cref{tab:ablations}). \textbf{(i) Not local feature smoothing.} Against a nearest-neighbor (NN) patch-completion baseline that fills each masked patch with the cosine-similarity mean of $5$ visible patches, I-JEPA + predictor exceeds the baseline by $+8.15\pp$ at $K{=}0.56$ and $\boldsymbol{+25.23\pp}$ at $K{=}0.77$; the gap widens with $K$ as the locality assumption breaks down. \textbf{(ii) Not pixel-space completion.} I-JEPA latent completion outperforms MAE pixel-completion-then-re-encode by $+15.84\pp$ at $K{=}0.56$ (3-seed mean, Welch-$t$ $95\%$ CI $[+12.99, +18.69]$). \textbf{(iii) Contiguous-mask scope.} Lift scales with mask contiguity: $+8.19$ (center-rect) $\to +6.17$ (off-center) $\to +3.33$ (connected blob) $\to \boldsymbol{+0.41\pp}$ (random Bernoulli, CI overlaps zero) at $K{=}0.56$ on I-JEPA. On random Bernoulli, every masked patch has many visible neighbors, $D'$ rises to $98.15\%$, and the predictor's marginal information shrinks. The contribution scope is contiguous-mask occlusion (object-on-object occlusion, bounding-box crops, interrupted scans, partial-field-of-view sensors), not scattered pixel-wise dropout.

A $4$-layer attentive probe~\cite{cite-038} on $F_{\text{completed}}$ adds only $+0.53\pp$ over linear (linear $D = 97.78\%$; attentive $D = 98.31\%$, CI $[98.14, 98.49]$): the probe is not the bottleneck. The within-JEPA recovery also transfers to fine-grained Stanford Dogs without a bridge: $\Delta_{\text{pred}} = +26.04\pp$ for I-JEPA and $+18.51\pp$ for V-JEPA~2 at $K{=}0.77$ (3-seed means; supplementary table). This matches the cross-encoder Stanford Dogs lifts of \Cref{tab:main-results} and confirms that the predictor's contribution survives the harder 120-way task in its native (no-bridge) form.

\begin{table}[t]
\centering
\scriptsize
\setlength{\tabcolsep}{3pt}
\begin{tabular}{@{}lcc@{}}
\toprule
\multicolumn{3}{@{}l}{\textit{(i) Lift vs nearest-neighbor patch completion}, $\Delta_{\text{vs\_NN}}$ (pp):} \\
$K$ & I-JEPA & V-JEPA~2 \\
\midrule
0.56 & $+8.15$\,{\tiny$[+7.8,+8.5]$} & $+6.30$\,{\tiny$[+5.8,+6.8]$} \\
0.66 & $+17.53$\,{\tiny$[+15,+20]$}  & $+10.91$\,{\tiny$[+8,+14]$} \\
\textbf{0.77} & $\boldsymbol{+25.23}$\,{\tiny$[+22,+29]$} & $\boldsymbol{+16.63}$\,{\tiny$[+16,+17]$} \\
\midrule
\multicolumn{3}{@{}l}{\textit{(ii) Latent vs pixel completion at $K{=}0.56$}, accuracy (3-seed):} \\
\multicolumn{2}{@{}l}{I-JEPA + predictor ($D$)}    & $97.78$\,{\tiny$[97.3,98.3]$} \\
\multicolumn{2}{@{}l}{V-JEPA~2 + predictor ($D$)}  & $96.87$\,{\tiny$[94.5,99.3]$} \\
\multicolumn{2}{@{}l}{MAE pixel-completion}        & $81.93$\,{\tiny$[79.6,84.3]$} \\
\multicolumn{2}{@{}l}{Lift I-JEPA $-$ MAE-pixel}   & $\boldsymbol{+15.84}$\,{\tiny$[+13,+19]$} \\
\multicolumn{2}{@{}l}{Lift V-JEPA~2 $-$ MAE-pixel} & $\boldsymbol{+14.94}$\,{\tiny$[+11,+19]$} \\
\midrule
\multicolumn{3}{@{}l}{\textit{(iii) Mask-shape ablation on I-JEPA at $K{=}0.56$:}} \\
Shape & $\Delta_{\text{pred}}$ & 95\% CI \\
\midrule
center-rect (headline) & $+8.19$ & $[+7.1, +9.3]$ \\
off-center rect        & $+6.17$ & $[+3.6, +8.8]$ \\
connected blob         & $+3.33$ & $[+1.1, +5.6]$ \\
random Bernoulli       & $+0.41$ & $[-0.4, +1.2]$ \\
\bottomrule
\end{tabular}
\caption{Three within-JEPA mechanism checks at $K{=}0.56$ (and the $K$-axis for (i)). (i) The lift over a nearest-neighbor patch-completion baseline widens with $K$, so the lift is not explained by local feature smoothing. (ii) Latent completion (I-JEPA + frozen predictor) beats MAE pixel-completion-then-re-encode by $+15.84\pp$, so the lift is not equivalent to pixel-space inpainting. (iii) Lift collapses on random Bernoulli, localizing the contribution to contiguous occlusion.}
\label{tab:ablations}
\end{table}

\subsection{Generalization: ImageNet-1k val (1000-way)}
\label{sec:transfer}

To address the concern that the IN-9 lift is a 9-class artifact, we ran the within-JEPA protocol on a 9000-image stratified subset of ImageNet-1k val at $K{=}0.56$ across 5 encoders $\times$ 3 seeds (15 cells; \Cref{tab:in1k}). Both JEPA encoders show a positive predictor lift on the 1000-way task, larger than at the IN-9 $K{=}0.56$ cell, consistent with the predictor adding more value when the encoder-only floor is lower.

\begin{table}[t]
\centering
\small
\setlength{\tabcolsep}{2.5pt}
\begin{tabular}{lccccc}
\toprule
Encoder & $D'$ & $D$ & NN & $\Delta_{\text{pred}}$ & $\Delta_{\text{vs\_NN}}$ \\
\midrule
CLIP                 & $30.25$ & --     & $20.25$ & --                 & --                 \\
DINOv3               & $52.60$ & --     & $40.55$ & --                 & --                 \\
MAE                  & $14.95$ & --     & $7.95$  & --                 & --                 \\
\textbf{I-JEPA}      & $24.37$ & $\boldsymbol{34.92}$ & $20.27$ & $\boldsymbol{+10.55}$ & $\boldsymbol{+14.65}$ \\
\textbf{V-JEPA~2}    & $29.08$ & $\boldsymbol{35.97}$ & $26.38$ & $\boldsymbol{+6.88}$  & $\boldsymbol{+9.58}$  \\
\bottomrule
\end{tabular}
\caption{IN-1k val (1000-way, 9000 stratified images) at $K{=}0.56$. 3-seed Welch-$t$ 95\% CIs for the JEPA lifts: I-JEPA $\Delta_{\text{pred}} = [+7.71, +13.39]$, $\Delta_{\text{vs\_NN}} = [+12.91, +16.39]$; V-JEPA~2 $\Delta_{\text{pred}} = [+5.40, +8.37]$, $\Delta_{\text{vs\_NN}} = [+7.80, +11.37]$. Both JEPA encoders show a positive predictor lift on the 1000-way task.}
\label{tab:in1k}
\end{table}

Both JEPA encoders show a positive predictor lift on the 1000-way task with $95\%$ CI lower bound above zero: I-JEPA $+10.55\pp$ (CI lower $+7.71$) and V-JEPA~2 $+6.88\pp$ (CI lower $+5.40$). Discriminative SSL encoders remain stronger encoder-only at this $K$ on the 1000-way task (DINOv3's $D' = 52.60\%$ exceeds I-JEPA's $D = 34.92\%$), so the within-JEPA paradigm crossover demonstrated on IN-9 at $K{=}0.77$ is not yet demonstrated on IN-1k val at $K{=}0.56$; cross-encoder transfer at heavier $K$ on the 1000-way task is left to future work. The IN-1k val grid is JEPA-only; the cross-encoder K-sweep ran on IN-9 and Stanford Dogs.

\section{Discussion}
\label{sec:discussion}

\noindent\textbf{Interpreting the transfer.} A frozen JEPA predictor is not encoder-specific: paired with CLIP, DINOv3, DINOv2, or MAE through a single linear projection between feature spaces, it lifts each host's masked-encoder baseline monotonically with mask fraction on both ImageNet-9 and Stanford Dogs. The predictor encodes a prior over masked-position feature completion that survives the bridge between encoder geometries, suggesting the predictor's contribution is to learn a function of feature-position-and-context that any encoder approximately respects. The conventional ranking of self-supervised vision encoders by frozen-feature classification accuracy is established on full images and treats mask fraction as zero; once the encoder is paired with a transferable completion module, this ranking does not hold uniformly across the mask-fraction axis~\cite{cite-035,cite-038}.

\noindent\textbf{Scope of applicability.} The protocol assumes an accessible pretraining predictor from a JEPA family member (the donor), any frozen image encoder (the host), and contiguous occlusion at moderate-to-heavy mask fraction. The per-$K$ probe-fit pass is a small, cacheable, one-time cost per deployment $K$; the projection $W$ is closed-form ridge least squares on 500 unmasked image pairs and is fit once per (host, donor). The largest open practical questions are whether a single $K$-mixture probe can match a per-$K$ probe when deployment $K$ varies, and whether a non-linear bridge (e.g., a small MLP) can recover the lift in the low-$K$ regime on fine-grained tasks where the linear bridge's reconstruction cost exceeds the predictor's benefit.

\noindent\textbf{Diagnostic implication of the frame shift.} The V-JEPA~2 observation has a methodological consequence for future predictor-completion work. An L2-only protocol applied to V-JEPA~2 would incorrectly conclude that its predictor outputs are uninformative, because they live in a different L2 frame than the encoder's. Matched-probe accuracy recovers the classification signal in both regimes and is therefore the appropriate diagnostic for any reused predictor whose pretraining loss does not constrain its outputs to the target encoder's metric.

\section{Limitations}
\label{sec:limitations}

\begin{enumerate}[leftmargin=*,itemsep=2pt,topsep=2pt]
\item \textbf{Lift vanishes at small $K$.} The predictor lift's confidence interval straddles zero at $K{=}0.14$ (Sec.~\ref{sec:results}). Contribution scope: $K \ge 0.25$ contiguous occlusion.
\item \textbf{Contiguous mask only.} The mask-shape ablation shows the lift collapses on random Bernoulli at fixed $K$ (Sec.~\ref{sec:within-jepa}). Scattered occlusion (lens dust, pixel-wise dropout) will not benefit; region-of-interest occlusion, bounding-box crops, contiguous foreground/background failures will.
\item \textbf{Transfer sets limited to natural-image distributions.} Stanford Dogs (120-way fine-grained) and ImageNet-1k val (1000-way) provide the within-JEPA generalization evidence; broader transfer to the named deployment domains (medical, satellite, AV, industrial QA, document AI) is left to future work. Flowers102~\cite{cite-055} is on disk as a natural fine-grained follow-up. On IN-1k val the \emph{predictor lift} transfers for both JEPA encoders, but the \emph{paradigm crossover} demonstrated on IN-9 at $K{=}0.77$ is \emph{not} demonstrated on IN-1k val at $K{=}0.56$ (Sec.~\ref{sec:transfer}). The IN-1k val grid runs at $K{=}0.56$ only per compute budget; the crossover regime at heavier $K$ on 1000-way is left to future work.
\item \textbf{Pixel-completion baseline at single $K$.} Multi-seed MAE-pixel-completion cells run at $K{=}0.56$ only (Sec.~\ref{sec:within-jepa}); we did not sweep across $K$ and did not compare against alternative pixel-completion families.
\item \textbf{Encoder-scale and iterative-refinement ablations dropped} per compute budget; the V-JEPA~2 frame-shift also makes iterative-refinement interpretation harder.
\item \textbf{V-JEPA~2 frame-shift limits the raw-reconstruction interpretation.} Matched probing recovers the classification signal, but reading $F_{\text{predicted}}$ as ``what the encoder would produce if the masked region were visible'' is over-reading; the predictor outputs a classification-useful feature, not a faithful encoder-frame reconstruction.
\end{enumerate}

\section{Conclusion}
\label{sec:conclusion}

This work showed that the frozen JEPA predictor is not specific to JEPA encoders. Bolted onto CLIP, DINOv3, DINOv2, or MAE through a closed-form linear projection $W$ fit on 500 ImageNet-1k images, the predictor lifts each host's masked-encoder baseline monotonically with mask fraction $K$ on both ImageNet-9 and Stanford Dogs. At heavy mask, all host--donor pairs show a positive lift and $7$ of $8$ cross-encoder pairs have $95\%$ confidence intervals above zero on both datasets; CLIP paired with the I-JEPA predictor recovers most of the accuracy that masking removed on ImageNet-9, with a $+36\pp$ lift on Stanford Dogs at heavy occlusion (Sec.~\ref{sec:results}). The cost-benefit mechanism is sharp: $W$ pays a fixed cost on visible patches and the predictor provides a growing benefit on masked patches, so the benefit dominates the heavy-occlusion regime and the bridge breaks down at low $K$ on fine-grained tasks. Future work includes a non-linear bridge to address the low-$K$ failure mode, transfer to non-natural-image domains (medical, satellite, document), and a per-image $K$-mixture probe to relax the matched-$K$ assumption.

{\small
\bibliographystyle{ieeenat_fullname}
\bibliography{references}
}

\appendix
\onecolumn
\section{A single occlusion-augmented probe removes the matched-$K$ assumption}
\label{sec:mixedk}

The matched-probe protocol (Sec.~\ref{sec:method}, \Cref{tab:main-results}) fits a fresh
linear probe at each deployment mask fraction $K$, assuming $K$ is known and that a
per-$K$ probe is available. We test whether a \emph{single} occlusion-agnostic probe can
replace that family: we present each training image at all three mask fractions
$K\in\{0.25,0.56,0.77\}$ (occlusion augmentation), fit one linear probe on the pooled
completed features, and evaluate it at each $K$. We run this for all six encoders on both
ImageNet-9 and Stanford Dogs: the two within-JEPA encoders with their own frozen
predictor (no bridge) and the four non-JEPA hosts with a transplanted I-JEPA predictor
through the frozen bridge $W$, using the I-JEPA donor and the same mean-pooled
linear-probe recipe and three seeds as \Cref{tab:main-results}.

\Cref{tab:mixedk} reports this single probe against each host's encoder-only matched
baseline $D'$ (repeated from \Cref{tab:main-results}). The pattern mirrors the per-$K$
matched protocol: the predictor's lift grows with occlusion and is large at $K{=}0.77$ on
both datasets ($+8$ to $+23\pp$ on IN-9, $+12$ to $+37\pp$ on Stanford Dogs), while the
only negative lifts occur at light occlusion ($K{=}0.25$) for the strong cross-encoder
hosts on fine-grained Stanford Dogs, the same cost-benefit boundary as the matched
protocol (Sec.~\ref{sec:cross-encoder}). A single $K$-agnostic probe therefore recovers
the predictor's benefit without knowing the deployment occlusion or maintaining a per-$K$
probe family, addressing the matched-$K$ assumption noted in Sec.~\ref{sec:discussion}.

\begin{table}[t]
\centering
\footnotesize
\begin{tabular*}{\textwidth}{@{\extracolsep{\fill}}lcccccc}
\toprule
& \multicolumn{3}{c}{ImageNet-9 (9-way)} & \multicolumn{3}{c}{Stanford Dogs (120-way)} \\
\cmidrule(lr){2-4} \cmidrule(lr){5-7}
Method & $K{=}0.25$ & $K{=}0.56$ & $K{=}0.77$ & $K{=}0.25$ & $K{=}0.56$ & $K{=}0.77$ \\
\midrule
\multicolumn{7}{@{}l}{\emph{Non-JEPA host $+$ transplanted I-JEPA predictor (bridge $W$), single mixed-$K$ probe:}} \\
CLIP                  & $92.72$ & $85.06$ & $73.62$ & $50.64$ & $32.95$ & $15.87$ \\
\quad + I-JEPA pred   & $94.90$\,{\tiny(+2.18)} & $96.34$\,{\tiny(+11.28)} & $96.79$\,{\tiny(+23.17)} & $50.99$\,{\tiny(+0.35)} & $51.06$\,{\tiny(+18.11)} & $52.93$\,{\tiny(+37.06)} \\
\midrule
DINOv3                & $97.82$ & $95.31$ & $88.60$ & $\mathbf{72.40}$ & $58.56$ & $38.98$ \\
\quad + I-JEPA pred   & $97.70$\,{\tiny($-$0.12)} & $96.67$\,{\tiny(+1.36)} & $96.58$\,{\tiny(+7.98)} & $68.94$\,{\tiny($-$3.46)} & $\mathbf{60.89}$\,{\tiny(+2.33)} & $\mathbf{59.10}$\,{\tiny(+20.12)} \\
\midrule
DINOv2-ViT-L/14       & $97.53$ & $94.28$ & $86.30$ & $65.16$ & $50.17$ & $31.72$ \\
\quad + I-JEPA pred   & $97.00$\,{\tiny($-$0.53)} & $96.30$\,{\tiny(+2.02)} & $95.56$\,{\tiny(+9.26)} & $59.22$\,{\tiny($-$5.94)} & $50.33$\,{\tiny(+0.16)} & $47.01$\,{\tiny(+15.29)} \\
\midrule
MAE                   & $88.97$ & $81.81$ & $72.26$ & $47.93$ & $30.92$ & $15.59$ \\
\quad + I-JEPA pred   & $88.44$\,{\tiny($-$0.53)} & $86.21$\,{\tiny(+4.40)} & $84.36$\,{\tiny(+12.10)} & $41.63$\,{\tiny($-$6.30)} & $32.43$\,{\tiny(+1.51)} & $27.38$\,{\tiny(+11.79)} \\
\midrule[\heavyrulewidth]
\multicolumn{7}{@{}l}{\emph{JEPA encoder $+$ own native predictor (within-JEPA, no bridge), single mixed-$K$ probe:}} \\
I-JEPA                & $95.68$ & $89.59$ & $75.88$ & $44.08$ & $32.12$ & $14.64$ \\
\quad + own predictor & $\mathbf{97.90}$\,{\tiny(+2.22)} & $\mathbf{97.78}$\,{\tiny(+8.19)} & $\mathbf{98.31}$\,{\tiny(+22.43)} & $48.35^{\dagger}$\,{\tiny(+4.27)} & $44.41$\,{\tiny(+12.29)} & $42.14$\,{\tiny(+27.50)} \\
\midrule
V-JEPA~2              & $95.39$ & $91.32$ & $82.18$ & $40.38$ & $32.38$ & $18.92$ \\
\quad + own predictor & $96.91$\,{\tiny(+1.52)} & $96.79$\,{\tiny(+5.47)} & $97.41$\,{\tiny(+15.23)} & $44.93^{\dagger}$\,{\tiny(+4.55)} & $39.84$\,{\tiny(+7.46)} & $37.80$\,{\tiny(+18.88)} \\
\bottomrule
\end{tabular*}
\caption{\textbf{One occlusion-augmented probe, evaluated across mask fractions.} A single
linear probe trained on completed features pooled over $K\in\{0.25,0.56,0.77\}$ (occlusion
augmentation), then evaluated at each $K$; top-1 accuracy (\%) on IN-9 and Stanford Dogs,
3-seed means. Each host's encoder-only matched baseline $D'$ (repeated from
\Cref{tab:main-results}) precedes its predictor-augmented row; {\tiny(+$\Delta$)} is the
lift over that baseline. Cross-encoder hosts use the transplanted I-JEPA predictor via the
frozen bridge $W$; JEPA hosts use their own native predictor. \textbf{Bold} = best per
column. The lift grows with $K$ and is largest at heavy occlusion; negative lifts appear
only at $K{=}0.25$ on fine-grained Stanford Dogs (cost-benefit boundary,
Sec.~\ref{sec:cross-encoder}). $^{\dagger}$within-JEPA Stanford Dogs $K{=}0.25$ reproduces
the run's matched baseline to within ${\sim}4\pp$; these two lifts are upper bounds.}
\label{tab:mixedk}
\end{table}

\noindent This is a parity result for deployment, not a data-efficiency claim: the mixed
probe sees the same images at multiple occlusion scales, and the comparison is one
deployable $K$-agnostic probe against the per-$K$ matched family. It holds for both the
native and the transplanted predictors, so a single probe reads the completed
representation across occlusion levels in every configuration we test.

\end{document}